\documentclass[10pt,twocolumn,letterpaper]{article}

\usepackage[pagenumbers]{iccv} 

\usepackage{amsmath}
\usepackage{amssymb}
\usepackage{graphicx}
\usepackage{booktabs} 
\usepackage{pgfplots} 
\pgfplotsset{compat=1.18} 
\usepackage{float} 
\usepackage{subcaption}
\usepackage{caption}
\expandafter\def\csname ver@subfig.sty\endcsname{}
\usepackage{xcolor} 
\usepackage{enumitem} 
\usepackage{geometry} 
\usepackage{cuted}
\usepackage{multicol}

\usepackage[most]{tcolorbox}
\usepackage{lmodern}

\definecolor{purplebg}{HTML}{2C2F54}
\definecolor{boxbg}{HTML}{1f2937}
\definecolor{titleviolet}{HTML}{BA91FF}
\definecolor{highlightpurple}{HTML}{AF78FF}
\definecolor{bodytext}{HTML}{d0d4da}
\definecolor{white}{HTML}{FFFFFF}
\definecolor{purple1}{HTML}{d8b4fe}
\definecolor{purple2}{HTML}{c084fc}

\definecolor{purple}{RGB}{192,128,255}
\definecolor{darkblue}{RGB}{20,30,48}
\definecolor{amber}{RGB}{255,204,0}
\definecolor{skyblue}{RGB}{102,178,255}
\definecolor{lightgreen}{RGB}{128,255,170}
\definecolor{salmonred}{RGB}{255,102,102}
\definecolor{whitetext}{RGB}{255,255,255}
\definecolor{background}{RGB}{22,27,51}

\tcbset{
  colframe=purple!60,
  colback=darkblue,
  coltext=white,
  boxrule=0.8pt,
  arc=2mm,
  top=1mm,
  bottom=1mm,
  left=2mm,
  right=2mm,
  fonttitle=\bfseries\small\sffamily,
  enhanced,
}

\newtcolorbox{stagebox}[1][]{
  colback=boxbg,
  colframe=highlightpurple,
  coltitle=white,
  coltext=bodytext,
  boxrule=0.8pt,
  arc=2pt,
  fonttitle=\bfseries\small\sffamily, 
  fontupper=\sffamily\scriptsize,          
  title=#1,
  enhanced,
  halign=left,
  breakable
}

\newtcolorbox{stagebox_2}[1][]{
  colback=boxbg,
  colframe=highlightpurple,
  coltext=bodytext,
  boxrule=0.8pt,
  arc=2pt,
  fontupper=\sffamily\scriptsize, 
  enhanced,
  halign=left,
  breakable
}

\definecolor{iccvblue}{rgb}{0.21,0.49,0.74}
\usepackage[pagebackref,breaklinks,colorlinks,allcolors=iccvblue]{hyperref}

\def\paperID{1} 
\def\confName{ICCV}
\def\confYear{2025}

\title{Aether Weaver: Multimodal Affective Narrative Co-Generation with Dynamic Scene Graphs}

\author{Saeed Ghorbani\\
{\tt\small theghorbani@gmail.com}
}

\begin{document}
\maketitle
\begin{abstract}
We introduce Aether Weaver, a novel, integrated framework for multimodal narrative co-generation that overcomes limitations of sequential text-to-visual pipelines. Our system concurrently synthesizes textual narratives, dynamic scene graph representations, visual scenes, and affective soundscapes, driven by a tightly integrated, co-generation mechanism. At its core, the Narrator, a large language model, generates narrative text and multimodal prompts, while the Director acts as a dynamic scene graph manager, and analyzes the text to build and maintain a structured representation of the story's world, ensuring spatio-temporal and relational consistency for visual rendering and subsequent narrative generation. Additionally, a Narrative Arc Controller guides the high-level story structure, influencing multimodal affective consistency, further complemented by an Affective Tone Mapper that ensures congruent emotional expression across all modalities. Through qualitative evaluations on a diverse set of narrative prompts encompassing various genres, we demonstrate that Aether Weaver significantly enhances narrative depth, visual fidelity, and emotional resonance compared to cascaded baseline approaches. This integrated framework provides a robust platform for rapid creative prototyping and immersive storytelling experiences.
\end{abstract}    
\section{Introduction}
\label{sec:intro}

The evolving landscape of digital media, encompassing interactive games, virtual reality, and automated content creation, demands increasingly sophisticated narrative generation capabilities, and while Large Language Models have demonstrated remarkable prowess in generating coherent textual narratives, their outputs often exist in isolation, lacking intrinsic visual and auditory context.  Conventional AI storytelling pipelines typically adopt a cascaded approach: generating text first, followed by separate, often disconnected, visual and auditory synthesis modules which generate multimodal story in a single run without intermediate user engagement \cite{kim2023multi, yang2024seed, arif2024art, doh2025exploratory}. This sequential processing frequently leads to critical shortcomings such as spatio-temporal inconsistencies, misaligned emotional tones, and a semantic disconnect between the narrative's evolving state and its multimodal manifestations.  For instance, a character's emotional shift or the sudden appearance of a critical object might not be accurately or coherently reflected in subsequent visual or auditory outputs, diminishing the overall immersive experience. 

This paper addresses the fundamental challenge of holistic narrative creation by proposing Aether Weaver, a novel, integrated framework for multimodal storytelling.  Unlike traditional sequential methods, Aether Weaver concurrently synthesizes textual narratives with their corresponding dynamic scene graph representations, directly generated visual scenes, and affective soundscapes.  Our core innovation lies in the dynamic scene graph manager (Director), which actively tracks and updates the spatial and relational state of entities within the narrative, ensuring unparalleled visual coherence and continuity across narrative segments.  This is complemented by a Narrative Arc Controller, providing high-level structural guidance for textual progression and multimodal affective trajectory,  and an Affective Tone Mapper, ensuring consistent emotional resonance across all generated modalities. 

Our contributions are summarized as follows:
\begin{itemize}[noitemsep,topsep=0pt]
    \item {A novel, integrated architecture that orchestrates simultaneous text, dynamic scene graph, direct visual scene generation, and affective soundscape co-generation, moving beyond conventional cascaded pipelines.}
    \item {The development of a dynamic scene-graph manager (Director) that maintains real-time visual coherence by tracking complex entity relationships and attributes throughout the narrative's progression. }
    \item {The design of a Narrative Arc Controller that guides the generative process according to predefined story structures, ensuring high-level narrative and multimodal affective consistency, and an Affective Tone Mapper that ensures congruent emotional expression across all modalities.}
\end{itemize}
\section{Related Work}
The field of AI-driven content generation is multidisciplinary, drawing from advancements in natural language processing, computer vision, and affective computing. Our work builds upon and extends several key areas.

\subsection{Text-to-Story Generation}
Recent breakthroughs in Large Language Models have enabled the generation of remarkably fluent and contextually relevant narratives. These models excel at maintaining character voice and plot points over short to medium narrative lengths. For instance, recent evaluations highlight LLMs' improved capabilities in generating creative short fiction, though they still show limitations in areas such as novelty, surprise, and diversity compared to human writers \cite{ismayilzada2024evaluating}. 

Furthermore, discussions on the ability of LLMs to engage with complex metafictional prompts indicate their evolving stylistic prowess \cite{scaleai2025llms}. However, a significant challenge in LLM-driven storytelling, especially for long-form narratives, remains to maintain global coherence and consistency across extended plotlines \cite{gurung2025learningreasonlongformstory}. Additionally, while they can describe visual scenes, they do not inherently generate structured representations suitable for direct visual synthesis, nor do they guarantee consistent entity tracking or spatio-temporal coherence across prolonged narratives without explicit external mechanisms. Approaches such as reasoning over condensed information or using hand-designed prompting techniques are being explored to address the long-form challenge \cite{gurung2025learningreasonlongformstory}, underscoring the need for more robust control mechanisms beyond simple prompting.

\subsection{Multimodal Content Synthesis}
Research in this domain focuses on translating textual narratives into visual layouts or concrete scenes, and more recently, into integrated multimodal experiences. Early methods often relied on rule-based systems or template matching \cite{zhang2022survey, wu2025icm}. More recent deep learning approaches have utilized techniques such as semantic parsing to extract scene elements, which are then fed into image generation models (e.g., GANs, diffusion models) to render visuals \cite{wu2025icm, zhang2025oasis}.

A significant advancement in this area is the development of Multimodal Large Language Models (MLLMs). For example, systems like SEED-Story propose using MLLMs with specialized attention mechanisms to generate multimodal long stories, encompassing both text and images, and often introduce new datasets to train and benchmark these integrated approaches \cite{yang2024seed}. Despite these innovations, a common limitation of many prior systems is their sequential operation: text is generated first, then parsed for visual cues, and finally, images are rendered. This decoupled process often leads to inconsistencies when an entity's state changes between narrative segments (e.g., an object appearing or disappearing without justification), or when complex inter-object relationships need to be maintained across a sequence of visuals. Furthermore, the generated visuals are typically static snapshots rather than components of a dynamically evolving scene, and the integration of audio is often an afterthought or relies on separate modules. Our work aims to overcome these limitations by fostering true co-generation across modalities.

\subsection{Affective Computing in Narrative}
This field explores the recognition, analysis, and generation of emotions within textual and multimodal narratives. Emotion-aware LLMs can generate text that reflects specific moods or character emotions \cite{zhang2024self, schlegel2025large}. In multimodal contexts, research has focused on synthesizing speech with emotional prosody or generating music that matches narrative moods \cite{triantafyllopoulos2023overview}.

However, integrating affective states dynamically and consistently across a full narrative, especially in a co-generative framework, remains an active area of research. Although some platforms explore AI-driven emotional reflection through narratives to scaffold emotional literacy \cite{ghosh2025narrative}, and discussions highlight current gaps in AI's ability to truly replicate the depth of human emotional experiences such as irony, humor, or empathy \cite{digitalcontentnext2024ais}, these underscore the need for more robust and nuanced affective control. Our work extends this by integrating affective control directly into the \textit{co-generation} loop of text, visual scenes, and soundscapes, ensuring that the emotional arc of the story is consistent and dynamically expressed across all modalities from the outset, rather than being an overlaid effect.

\subsection{Scene Graph Representations}
Scene graphs have proven invaluable in computer vision for representing objects, their attributes, and their relationships within an image or video \cite{johnson2018image}. They provide a structured, symbolic representation that facilitates complex reasoning and generation tasks. While primarily used for \textit{understanding} existing visual content, there is also an emerging work on \textit{generating} images from scene graphs \cite{krishna2017visual, johnson2018image}.

Crucially, the application of knowledge graphs, including scene graphs, extends to improving consistency and logical reasoning in story generation. Research has shown that leveraging commonsense knowledge graphs and axioms can lead to more sensible story endings and improve the underlying narrative structures \cite{ilievski2021story}. Furthermore, approaches to controllable text generation using external knowledge bases demonstrate the power of such structured data to guide narrative creation \cite{xu2020megatron}. Recent work also explores controllable logical hypothesis generation using knowledge graphs to enhance abductive reasoning, which can be applied to scenario planning in dynamic environments \cite{gao2025controllable}. Our work significantly leverages the power of scene graphs not just for static image generation but for dynamically tracking and updating the entire visual and conceptual world of a narrative as it unfolds. This dynamic aspect, coupled with its role in driving multimodal co-generation and ensuring spatio-temporal and relational coherence across a story's progression, distinguishes our approach from prior work by integrating the scene graph as a central, continuously updated truth source for the entire narrative.
\section{Aether Weaver Architecture}
\begin{figure*}[htbp]
\centering
\includegraphics[width=1.0\linewidth]{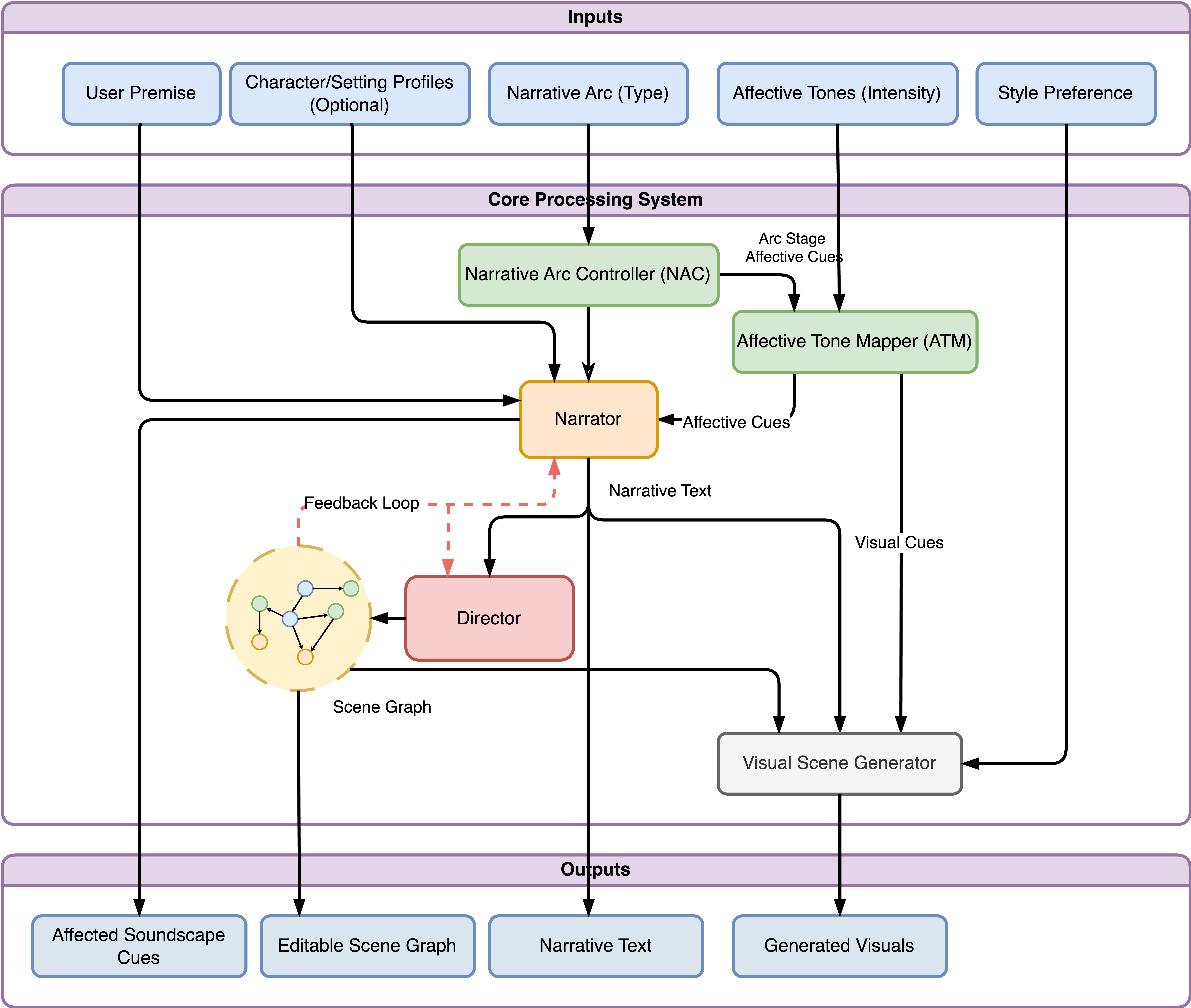}
\caption{Aether Weaver System Architecture. This block diagram illustrates the comprehensive flow of information within Aether Weaver. User inputs (Story Premise, Narrative Arc, Affective Tones, Character/Setting Profiles, Style Preference) feed into the system. The \textbf{Narrative Arc Controller (NAC)} and \textbf{Affective Tone Mapper (ATM)} provide high-level guidance to the \textbf{Narrator}. The Narrator generates textual narrative segments which are then processed by the \textbf{Director} designed as a Dynamic Scene Graph Manager. Finally, the system produces multimodal outputs: Narrative Text, a Dynamic Scene Graph, Generated Visuals (by \textbf{Visual Scene Generator}), and Directly Generated Soundscapes.}
\label{architecture}
\end{figure*}

Aether Weaver is designed as an end-to-end multimodal narrative generation system. Its architecture, depicted in \textbf{Figure \ref{architecture}}, integrates several key components that operate in concert to ensure coherence and affective consistency across textual, visual, and auditory modalities.

\subsection{Narrator}
At the core of Aether Weaver is a pre-trained large language model. This LLM serves as the primary generative engine, synthesizing textual narrative segments and affective soundscape cues based on a system of prompts and dynamic context. It receives detailed guidance from the Narrative Arc Controller (NAC) and Affective Tone Mapper (ATM), as well as critical feedback loops from Director.

\noindent\textbf{Prompt Engineering:} Our approach employs a multi-turn, iterative prompting strategy. The LLM's system instructions define its role as a multimodal storyteller. For each narrative segment, the prompt is dynamically constructed to include: (1) user inputs (user premise, character/setting profiles, visual and style preferences), (2) summary of story so far, (2) narrative arc type, current stage, and the objective from the NAC (e.g., ``introduce a new conflict'', ``escalate tension''), (3) desired affective cues from the ATM (e.g., ``Mystery: Focus on secrets, puzzles, and hidden truths''), and (4) the current state of the scene graph. This comprehensive prompting allows for fine-grained control over the generated text, ensuring it aligns with structural, emotional, and visual requirements. For each scene, the LLM then outputs a textual narrative segment and affective soundscape cues.

\subsection{Narrative Arc Controller (NAC)}
NAC is responsible for translating abstract narrative structures (e.g., ``Classic Arc'', ``Tragedy'') into concrete, actionable directives for the Narrator. It functions as a state machine, progressing through predefined narrative beats (Exposition, Rising Action, Climax, Falling Action, Resolution). Our implementation offers a selection of common narrative arcs, including Classic Arc, Tragedy, Hero's Journey, Rags to Riches, Quest, and Character Arc.

\noindent\textbf{Mechanism:} Based on the user's selected arc, the NAC internally defines a sequence of target states or ``plot points''. At each juncture, it generates a concise directive for the Narrator. For instance, transitioning from ``Exposition'' to ``Rising Action'' might involve directing the Narrator to ``introduce an inciting incident'' or ``reveal a new challenge''. The NAC also informs the Affective Tone Mapper about the expected emotional trajectory for the current arc phase, allowing for pre-emptive affective shifts. This top-down control ensures the generated narrative maintains a coherent structure, preventing aimless or meandering plots.

The NAC employs a ratio-based progression mechanism to guide the story's overall structure. This involves pre-defining the proportional length of each narrative stage relative to the total story duration. It tracks the current progress, triggering transitions to subsequent stages and issuing corresponding high-level objectives and affective cues to the generative modules as each predefined ratio boundary is met. While more sophisticated strategies like event-based triggers or narrator-based assessments are possible, the ratio-based approach proved effective and efficient for our implementation.

\subsection{Affective Tone Mapper (ATM)}
The ATM is responsible for translating abstract emotional intent into concrete, actionable directives for the generative components. It ensures emotional consistency and depth throughout a generated story by translating a user's desired emotional tones, each assigned an intensity (Low, Medium, or High), and optionally considering emotional cues from the current Narrative Arc stage, into practical guidance for Narrator and Visual Scene Generator. The selected intensity dictates how strongly and prominently each tone is expressed.

\noindent\textbf{Mechanism:} The ATM uses an internal knowledge base to interpret these emotional inputs and their intensities, dynamically adapting them based on narrative context if Narrative Arc Stage (NAC) influence is enabled (e.g., modulating ``High Joy'' to a lower effective intensity during a tragic story stage). Ultimately, the ATM produces three key outputs reflecting this intensity: a detailed affective directive for the Narrator, a list of soundscape cues, and a list of visual cues. By orchestrating these elements, the ATM acts as an emotional conductor, ensuring the story's text, visuals, and suggested sounds work harmoniously to evoke the intended feelings at the specified strength, thereby enhancing the overall immersive and emotionally resonant experience of Aether Weaver.

\begin{figure*}[htbp]
\centering
\includegraphics[width=0.7\linewidth]{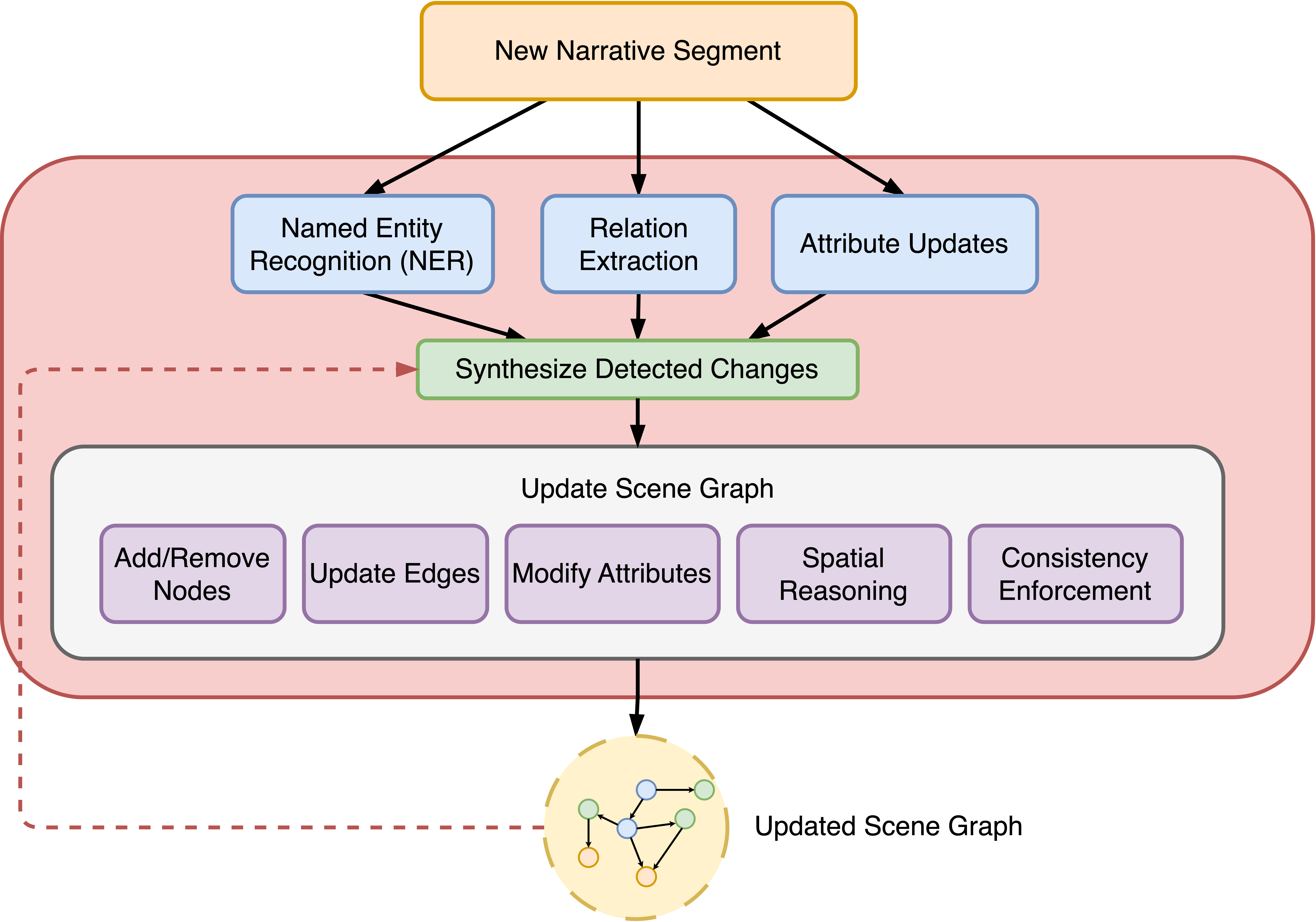}
\caption{Dynamics scene graph update by Director. 
}
\label{fig:dsgm}
\end{figure*}

\subsection{Director}
The Director is the cornerstone of our multimodal coherence, ensuring visual consistency and tracking the evolving state of the narrative world. It is designed as an LLM-based Dynamic Scene Graph Manager that operates in an iterative loop with the Narrator, as shown in \textbf{Figure \ref{fig:dsgm}}. 

\noindent\textbf{Definition of Scene Graph:} Our scene graph $\mathcal{G} = (\mathcal{V}, \mathcal{E}, \mathcal{A})$ is a directed graph where $\mathcal{V}$ is the set of nodes representing entities (e.g., characters, objects, locations), $\mathcal{E}$ is the set of directed edges representing relationships between entities (e.g., ``Elara $\xrightarrow{\text{holds}}$ datapad''), and $\mathcal{A}$ is the set of attributes associated with nodes or edges (e.g., ``Elara $\xrightarrow{\text{attribute}}$ weary'', ``datapad $\xrightarrow{\text{attribute}}$ flickering'').

\noindent\textbf{Dynamic Update Mechanism:} After each narrative segment is generated by the Narrator, the Director performs a number of tasks to translate unstructured natural language (the Narrator's generated text) into structured graph updates.
\textbf{Named Entity Recognition (NER)} sub-module identifies all new and existing characters, objects, and locations mentioned in the segment.
\textbf{Relation Extraction (RE)} sub-module extracts relationships (e.g., actions, spatial relations, causal links) between these entities. This often involves dependency parsing and semantic role labeling on the generated text.
\textbf{Attribute Updates} sub-module identifies changes in the properties or states of entities (e.g., ``The once gleaming artifact \textit{was now tarnished}''). Upon identifying required changes, the scene graph is updated by adding or removing nodes and edges, modifying attributes, and performing basic spatial reasoning.
\textbf{Spatial Reasoning} sub-module tracks relative positions (e.g., ``A is near B'', ``C is inside D'', ``D is on E''). This prevents entities from inexplicably teleporting or overlapping without narrative justification.
\textbf{Consistency Enforcement} sub-module performs basic logical checks to ensure updates are plausible. For instance, if an entity is removed from a scene, its associated relationships are also pruned. If a character is described as moving, their previous location attribute is updated. In cases of detected inconsistency, a flag can be raised or a corrective prompt sent back to the Narrator.

\subsection{Visual Scene Generator}
The \textbf{Visual Scene Generator} synthesizes visual scenes, drawing upon detailed, structured information from the Narrator, Director, and ATM modules.

\subsubsection{Mechanism}
This module generates scene images by first compiling all relevant information about the current moment in the story. This includes the narrative text, characters' appearances, a structured understanding of characters, the scene graph, user-defined visual style preferences, and the scene's mood (influenced by affective tones and the narrative arc). An LLM then synthesizes this rich collection of data into a single, detailed descriptive prompt which is then sent to an image generation model to create the final scene image. 

To ensure consistent and coherent visual representation of characters throughout the narrative, our system employs a multimodal character grounding technique. This approach first checks for a user-provided full-body portraits to establish a definitive visual anchor for each character. When such portraits are not available or when additional visual fidelity is needed, the system automatically generates a unique full-body image based on the textual description of character and selected stylistic parameters. The resulting character image, whether user-uploaded or generated, is then processed by a multimodal analysis model. This model extracts fine-grained descriptive features related to the character’s appearance, attire, and other distinguishing visual traits. These extracted attributes are used to automatically populate and enrich the character's profile, creating a stable visual-textual foundation. This foundation ensures high fidelity and consistency in the character’s appearance across all story elements and illustrations.

\subsection{User Editable Content}
Our design offers users two primary avenues for creative control: direct editing and plot twists. Users can directly edit a scene's narrative text or its associated scene graph. When a user saves a modified narrative, the Director module automatically re-analyzes the text and updates its internal graph representation, which can lead to the identification of new characters (and subsequent profile generation) or modifications to existing ones. Following this, new visual and soundscape cues are generated, reflecting the updated narrative and scene graph.

For dynamic story evolution, the Aether Weaver includes a Plot Twist feature. Users can select one of the LLM-generated plot twists or input a custom one. The chosen twist then serves as a directive for the Narrator, guiding the generation of the subsequent scene to incorporate this new element and significantly alter the story's direction. This feature empowers user-influenced narrative evolution and helps overcome creative impasses.
\section{Evaluation}
To evaluate Aether Weaver, we employed qualitative metrics derived from expert human analysis of its generated multimodal content.

\subsection{Experimental Setup}
We curated a diverse set of 10 narrative prompts, each designed to test specific aspects of multimodal coherence and narrative progression. These prompts encompassed various genres (e.g., Sci-Fi, Fantasy), character archetypes, initial settings, and complex plot hooks. For each prompt, we manually defined one to three key affective tones with varying intensities (e.g., ``Mystery: High'', ``Hope: Medium'', ``Loneliness: Low'') and selected a standard narrative arc (e.g., ``Classic Arc'', ``Tragedy'', ``Quest''). This ensures a robust test bed for the system's ability to handle diverse creative inputs and structural constraints.

 For comparison, we implemented a baseline system that operates sequentially similar to \cite{arif2024art}: a standard LLM (without our NAC or DSGM loops) generates the full narrative text first. Then, a separate module performs basic NER and feeds these extracted entities to a generic image prompt generator and a basic soundscape descriptor (e.g., ``ominous sounds'' if the text mentions ``ominous''). There's no dynamic scene graph or affective co-generation.

\subsection{Evaluation Metrics}
A panel of 10 expert human evaluators comprising professional writers, game designers, and AI researchers with experience in narrative systems assessed the generated outputs. Each metric was scored on a 5-point Likert scale (1=Poor, 5=Excellent).
Expert evaluators assessed narrative fluency,  multimodal coherence (alignment between visuals, sound, and narrative), affective alignment(consistency and evolution of emotional tone, and narrative arc Adherence.

\subsection{Results}
As shown in \textbf{Table \ref{tab:1}} our evaluation demonstrates that Aether Weaver significantly outperforms the cascaded baseline system across all key metrics, highlighting the profound benefits of our integrated co-generation approach and dynamic scene graph management.

\begin{table}[htbp]
    \small
    \centering
    \caption{Human Evaluation Results (Mean Likert Score, 1–5), comparing Aether Weaver (AW) against cascade baseline}
    \begin{tabular}{@{}lcc@{}}
        \toprule
        \textbf{Metric} & \textbf{AW} & \textbf{Baseline} \\
        \midrule
        Narrative Fluency            & \textbf{4.62} & 4.21 \\
        MM Coherence (Visual)        & \textbf{4.23} & 2.85 \\
        MM Coherence (Auditory)      & \textbf{4.76} & 4.10 \\
        Affective Alignment          & \textbf{4.67} & 3.28 \\
        Narrative Arc Adherence      & \textbf{4.30} & 3.45 \\
        \bottomrule
    \end{tabular}
    \label{tab:1}
\end{table}

\noindent\textbf{Narrative Fluency:} While both systems produced fluent and grammatically correct text, Aether Weaver, benefiting from the structured context provided by the NAC and Director feedback loops, exhibited slightly higher logical consistency and fewer instances of abrupt topic shifts or minor inconsistencies in character behavior. This led to a notable improvement in overall narrative flow and readability.

\noindent\textbf{Multimodal Coherence (Visual):} 
This metric showed the most dramatic improvement, with Aether Weaver scoring significantly higher than the baseline. By actively tracking entities and their relationships, it consistently generated visuals that reflected the evolving narrative. Evaluators noted that the baseline often forgot objects, rendered inconsistent poses or locations, and failed to update object states, resulting in a coherence gap. In contrast, Aether Weaver’s scene graph-driven visuals maintained stronger spatio-temporal and relational consistency.

\noindent\textbf{Multimodal Coherence (Auditory):} 
Aether Weaver's integration of the Affective Tone Mapper with the Director's scene awareness allowed it to generate soundscapes more aligned with both the story's emotional tone and unfolding events. For instance, a character's ``growing dread'' might trigger ``low, rising dissonance'' and ``exaggerated heartbeat sounds'', while the baseline typically offered generic cues like ``ominous music.''

\noindent\textbf{Affective Alignment:} Aether Weaver, through the combined efforts of the NAC and ATM, was able to not only initiate stories with the desired affective tone but also to smoothly transition and evolve these tones throughout the narrative, mirroring the selected arc. The baseline system, lacking explicit affective control beyond initial prompting, often exhibited flat emotional trajectories or abrupt, unmotivated emotional shifts.

\noindent\textbf{Narrative Arc Adherence:} Aether Weaver's NAC successfully guided the LLM to include key narrative beats (e.g., inciting incident, climax, resolution) at appropriate stages, leading to stories that felt more structurally complete and satisfying. The baseline, without this explicit structural control, frequently produced narratives that meandered, lacked a clear progression towards a resolution, or had poorly defined turning points.
\section{Conclusion and Future Work}
We presented Aether Weaver, a novel framework for multimodal affective narrative co-generation that integrates textual storytelling, dynamic scene graph management, visual scene generation, and affective soundscape synthesis. Leveraging a scene graph-based Director for visual coherence and a Narrative Arc Controller for structural guidance, Aether Weaver outperforms traditional cascaded pipelines in multimodal consistency, narrative fluency, and affective alignment. It provides an intuitive tool for rapidly prototyping emotionally rich and visually coherent stories. Several limitations remain, each suggesting clear paths forward. The current Director employs simplified relative positioning, limiting spatial reasoning in dynamic environments. Incorporating 3D scene representations and physics-based reasoning—potentially via integration with game engines could support more complex spatial interactions.
Implicit emotional shifts remain difficult to capture across modalities. Enriching character models with psychological depth and character-specific knowledge bases may improve nuanced emotional expression and behavioral consistency.
Maintaining long-form narrative coherence, especially with evolving plots and characters, still benefits from human oversight. Enhancing memory and narrative planning mechanisms could reduce this dependency while enabling scalable storytelling.
Scene-to-scene visual consistency is limited by de novo text-to-image generation. Image conditioning techniques or personalized generation such as DreamBooth~\cite{ruiz2023dreambooth} and textual inversion~\cite{gal2022image} could enable persistent character and setting representations.
Finally, while the system generates affective soundscapes, it lacks character speech. Incorporating expressive speech synthesis and dialogue generation—using techniques like voice cloning and emotional speech models~\cite{liu2023audioldm} would further enhance narrative immersion.

{
    \small
    \bibliographystyle{ieeenat_fullname}
    \bibliography{main}

\begin{thebibliography}{23}
\providecommand{\natexlab}[1]{#1}
\providecommand{\url}[1]{\texttt{#1}}
\expandafter\ifx\csname urlstyle\endcsname\relax
  \providecommand{\doi}[1]{doi: #1}\else
  \providecommand{\doi}{doi: \begingroup \urlstyle{rm}\Url}\fi

\bibitem[Arif et~al.(2024)Arif, Arif, Haroon, Khan, Raza, and Athar]{arif2024art}
Samee Arif, Taimoor Arif, Muhammad~Saad Haroon, Aamina~Jamal Khan, Agha~Ali Raza, and Awais Athar.
\newblock The art of storytelling: Multi-agent generative ai for dynamic multimodal narratives.
\newblock \emph{arXiv preprint arXiv:2409.11261}, 2024.

\bibitem[{Digital Content Next}(2024)]{digitalcontentnext2024ais}
{Digital Content Next}.
\newblock {AI's Impact on Storytelling: Can It Replicate Human Experiences?}
\newblock \url{https://digitalcontentnext.org/blog/2024/11/20/ais-impact-on-storytelling-can-it-replicate-human-experiences/}, 2024.

\bibitem[Doh et~al.(2025)Doh, Shi, Jain, Kim, and Ramani]{doh2025exploratory}
Hyungjun Doh, Jingyu Shi, Rahul Jain, Heesoo Kim, and Karthik Ramani.
\newblock An exploratory study on multi-modal generative ai in ar storytelling.
\newblock \emph{arXiv preprint arXiv:2505.15973}, 2025.

\bibitem[Gal et~al.(2022)Gal, Alaluf, Atzmon, Patashnik, Bermano, Chechik, and Cohen-Or]{gal2022image}
Rinon Gal, Yuval Alaluf, Yuval Atzmon, Or Patashnik, Amit~H Bermano, Gal Chechik, and Daniel Cohen-Or.
\newblock An image is worth one word: Personalizing text-to-image generation using textual inversion.
\newblock \emph{arXiv preprint arXiv:2208.01618}, 2022.

\bibitem[Gao et~al.(2025)Gao, Bai, Zheng, Sun, Zhang, Li, Song, and Fu]{gao2025controllable}
Yisen Gao, Jiaxin Bai, Tianshi Zheng, Qingyun Sun, Ziwei Zhang, Jianxin Li, Yangqiu Song, and Xingcheng Fu.
\newblock Controllable logical hypothesis generation for abductive reasoning in knowledge graphs.
\newblock \emph{arXiv preprint arXiv:2505.20948}, 2025.

\bibitem[Ghosh et~al.(2025)Ghosh, Chen, Alghabra, and Ho]{ghosh2025narrative}
Prerana Ghosh, Sherry Chen, Mohamed Alghabra, and Anthony Ho.
\newblock Narrative-centered emotional reflection: Scaffolding autonomous emotional literacy with ai.
\newblock \emph{arXiv preprint arXiv:2504.20342}, 2025.

\bibitem[Gurung and Lapata(2025)]{gurung2025learningreasonlongformstory}
Alexander Gurung and Mirella Lapata.
\newblock Learning to reason for long-form story generation, 2025.

\bibitem[Ilievski et~al.(2021)Ilievski, Pujara, and Zhang]{ilievski2021story}
Filip Ilievski, Jay Pujara, and Hanzhi Zhang.
\newblock Story generation with commonsense knowledge graphs and axioms.
\newblock In \emph{Workshop on Commonsense Reasoning and Knowledge Bases}, 2021.

\bibitem[Ismayilzada et~al.(2024)Ismayilzada, Stevenson, and van~der Plas]{ismayilzada2024evaluating}
Mete Ismayilzada, Claire Stevenson, and Lonneke van~der Plas.
\newblock Evaluating creative short story generation in humans and large language models.
\newblock \emph{arXiv preprint arXiv:2411.02316}, 2024.

\bibitem[Johnson et~al.(2018)Johnson, Gupta, and Fei-Fei]{johnson2018image}
Justin Johnson, Agrim Gupta, and Li Fei-Fei.
\newblock Image generation from scene graphs.
\newblock In \emph{Proceedings of the IEEE conference on computer vision and pattern recognition}, pages 1219--1228, 2018.

\bibitem[Kim et~al.(2023)Kim, Heo, Yu, and Nang]{kim2023multi}
Juntae Kim, Yoonseok Heo, Hogeon Yu, and Jongho Nang.
\newblock A multi-modal story generation framework with ai-driven storyline guidance.
\newblock \emph{Electronics}, 12\penalty0 (6):\penalty0 1289, 2023.

\bibitem[Krishna et~al.(2017)Krishna, Zhu, Groth, Johnson, Hata, Kravitz, Chen, Kalantidis, Li, Shamma, et~al.]{krishna2017visual}
Ranjay Krishna, Yuke Zhu, Oliver Groth, Justin Johnson, Kenji Hata, Joshua Kravitz, Stephanie Chen, Yannis Kalantidis, Li-Jia Li, David~A Shamma, et~al.
\newblock Visual genome: Connecting language and vision using crowdsourced dense image annotations.
\newblock \emph{International journal of computer vision}, 123:\penalty0 32--73, 2017.

\bibitem[Liu et~al.(2023)Liu, Chen, Yuan, Mei, Liu, Mandic, Wang, and Plumbley]{liu2023audioldm}
Haohe Liu, Zehua Chen, Yi Yuan, Xinhao Mei, Xubo Liu, Danilo Mandic, Wenwu Wang, and Mark~D Plumbley.
\newblock Audioldm: Text-to-audio generation with latent diffusion models.
\newblock \emph{arXiv preprint arXiv:2301.12503}, 2023.

\bibitem[Ruiz et~al.(2023)Ruiz, Li, Jampani, Pritch, Rubinstein, and Aberman]{ruiz2023dreambooth}
Nataniel Ruiz, Yuanzhen Li, Varun Jampani, Yael Pritch, Michael Rubinstein, and Kfir Aberman.
\newblock Dreambooth: Fine tuning text-to-image diffusion models for subject-driven generation.
\newblock In \emph{Proceedings of the IEEE/CVF conference on computer vision and pattern recognition}, pages 22500--22510, 2023.

\bibitem[{Scale AI Blog}(2025)]{scaleai2025llms}
{Scale AI Blog}.
\newblock {LLMs Are Getting Better at Generating Short Fiction}.
\newblock \url{https://scale.com/blog/llms-generating-fiction}, 2025.

\bibitem[Schlegel et~al.(2025)Schlegel, Sommer, and Mortillaro]{schlegel2025large}
Katja Schlegel, Nils~R Sommer, and Marcello Mortillaro.
\newblock Large language models are proficient in solving and creating emotional intelligence tests.
\newblock \emph{Communications Psychology}, 3\penalty0 (1):\penalty0 1--14, 2025.

\bibitem[Triantafyllopoulos et~al.(2023)Triantafyllopoulos, Schuller, {\.I}ymen, Sezgin, He, Yang, Tzirakis, Liu, Mertes, Andr{\'e}, et~al.]{triantafyllopoulos2023overview}
Andreas Triantafyllopoulos, Bj{\"o}rn~W Schuller, G{\"o}k{\c{c}}e {\.I}ymen, Metin Sezgin, Xiangheng He, Zijiang Yang, Panagiotis Tzirakis, Shuo Liu, Silvan Mertes, Elisabeth Andr{\'e}, et~al.
\newblock An overview of affective speech synthesis and conversion in the deep learning era.
\newblock \emph{Proceedings of the IEEE}, 111\penalty0 (10):\penalty0 1355--1381, 2023.

\bibitem[Wu et~al.(2025)Wu, Zhao, Cao, Xu, Jiang, Wang, Li, Hu, Qin, and Fu]{wu2025icm}
Mengyang Wu, Yuzhi Zhao, Jialun Cao, Mingjie Xu, Zhongming Jiang, Xuehui Wang, Qinbin Li, Guangneng Hu, Shengchao Qin, and Chi-Wing Fu.
\newblock Icm-assistant: Instruction-tuning multimodal large language models for rule-based explainable image content moderation.
\newblock In \emph{Proceedings of the AAAI Conference on Artificial Intelligence}, pages 8413--8422, 2025.

\bibitem[Xu et~al.(2020)Xu, Patwary, Shoeybi, Puri, Fung, Anandkumar, and Catanzaro]{xu2020megatron}
Peng Xu, Mostofa Patwary, Mohammad Shoeybi, Raul Puri, Pascale Fung, Anima Anandkumar, and Bryan Catanzaro.
\newblock Megatron-cntrl: Controllable story generation with external knowledge using large-scale language models.
\newblock \emph{arXiv preprint arXiv:2010.00840}, 2020.

\bibitem[Yang et~al.(2024)Yang, Ge, Li, Chen, Ge, Shan, and Chen]{yang2024seed}
Shuai Yang, Yuying Ge, Yang Li, Yukang Chen, Yixiao Ge, Ying Shan, and Yingcong Chen.
\newblock Seed-story: Multimodal long story generation with large language model.
\newblock \emph{arXiv preprint arXiv:2407.08683}, 2024.

\bibitem[Zhang et~al.(2025)Zhang, Cui, Zhao, and Yang]{zhang2025oasis}
Letian Zhang, Quan Cui, Bingchen Zhao, and Cheng Yang.
\newblock Oasis: One image is all you need for multimodal instruction data synthesis.
\newblock \emph{arXiv preprint arXiv:2503.08741}, 2025.

\bibitem[Zhang et~al.(2024)Zhang, Naradowsky, and Miyao]{zhang2024self}
Qiang Zhang, Jason Naradowsky, and Yusuke Miyao.
\newblock Self-emotion blended dialogue generation in social simulation agents.
\newblock \emph{arXiv preprint arXiv:2408.01633}, 2024.

\bibitem[Zhang et~al.(2022)Zhang, Li, Wei, Pan, and Deng]{zhang2022survey}
Ziqi Zhang, Zeyu Li, Kun Wei, Siduo Pan, and Cheng Deng.
\newblock A survey on multimodal-guided visual content synthesis.
\newblock \emph{Neurocomputing}, 497:\penalty0 110--128, 2022.

\end{thebibliography}
}
\clearpage
\section*{Supplementary Material}
\subsection*{Narrative Arc Controller Design}
The Narrative Arc Controller (NAC) directly guides the Narrator by providing the current stage name (e.g., ``climax'') and a specific goal directive (e.g., ``Reach the peak of conflict'') for each scene. This ensures the narrative aligns with the selected story structure. NAC also influences the Affective Tone Mapper (ATM) by always contributing its current stage's predefined affective cues to the ATM's processing. Additionally, the NAC's impact may become stronger: a dominant negative tone from the NAC stage (like ``Despair - High'') can reduce the intensity of any positive tones the user selected (e.g., ``Joy - High'' might become ``Joy - Medium''). The final instructions sent to the language model then reflect these combined and potentially modulated affective tones. Below is an example of narrative arc and its controlling parameters.
\begin{tcolorbox}[
    colback=boxbg,
    colframe=highlightpurple,
    coltext=bodytext,
    boxrule=1pt,
    arc=3pt,
    enhanced,
    title=Classic Arc,
    coltitle=white,
    fonttitle=\bfseries\normalsize\sffamily,
    breakable,
    halign=left,
]

\footnotesize\sffamily\textit{A fundamental plot structure involving rising action, climax, and resolution.}


\begin{stagebox}[Exposition, halign=left]
\textcolor{purple2}{\textbf{Goal:}} Introduce the main character(s), setting, and initial status quo.

\textcolor{purple2}{\textbf{Affective Cues:}} Peacefulness (Medium), Curiosity (Low)

\textcolor{purple2}{\textbf{Min. Story Progress for Stage:}} 5\%
\end{stagebox}





\begin{stagebox}[Rising Action, halign=left]
\textcolor{purple2}{\textbf{Goal:}} Develop conflicts, introduce obstacles, and build tension towards the climax. Protagonist faces escalating challenges.

\textcolor{purple2}{\textbf{Affective Cues:}} Suspense (High), Tension (High), Hope (Medium)

\textcolor{purple2}{\textbf{Keywords:}} struggle, fight, discovery, obstacle, plan, train

\textcolor{purple2}{\textbf{Min. Story Progress for Stage:}} 25\%
\end{stagebox}

\begin{stagebox}[Climax, halign=left]
\textcolor{purple2}{\textbf{Goal:}} Reach the peak of conflict and tension; the protagonist faces their greatest challenge and makes a decisive choice or action.

\textcolor{purple2}{\textbf{Affective Cues:}} Tension (High), Fear (Medium), Hope (Low)

\textcolor{purple2}{\textbf{Keywords:}} final confrontation, decisive action, all or nothing, showdown, turning point, revelation

\textcolor{purple2}{\textbf{Min. Story Progress for Stage:}} 65\%
\end{stagebox}

\begin{stagebox}[Falling Action, halign=left]
\textcolor{purple2}{\textbf{Goal:}} Show the immediate aftermath of the climax. Conflicts begin to resolve, tension decreases.

\textcolor{purple2}{\textbf{Affective Cues:}} Sadness (Low), Peacefulness (Medium)

\textcolor{purple2}{\textbf{Keywords:}} aftermath, consequences, healing, return, winding down

\textcolor{purple2}{\textbf{Min. Story Progress for Stage:}} 80\%
\end{stagebox}

\begin{stagebox}[Resolution, halign=left]
\textcolor{purple2}{\textbf{Goal:}} Tie up loose ends. The new status quo is established. Character arcs conclude.

\textcolor{purple2}{\textbf{Affective Cues:}} Peacefulness (High), Hope (Medium)

\textcolor{purple2}{\textbf{Keywords:}} resolved, new normal, ends, future, legacy

\textcolor{purple2}{\textbf{Min. Story Progress for Stage:}} 90\%
\end{stagebox}
\end{tcolorbox}
\vspace{1cm}
\subsection*{Affective Tone Mapper Design}
The Affective Tone Mapper (ATM) processes user-selected affective tones and, if enabled, cues from the current Narrative Arc (NAC) stage. It leverages a knowledge base to translate these tones into practical guidance. The ATM then outputs a detailed directive for the Narrator, explaining how to express tones and any modulations. Additionally, it provides lists of sound and visual cues to guide soundscape and image generation, ensuring affective consistency. The ATM also includes intensity-based keywords for more granular control. Below is an example of the knowledge base used for a specific affective tone.
\clearpage
\begin{tcolorbox}[
    colback=boxbg,         
    colframe=highlightpurple, 
    coltext=white,         
    boxrule=1pt,              
    arc=3pt,                  
    enhanced,                 
    title=Mystery,            
    coltitle=white,           
    fonttitle=\bfseries\normalsize\sffamily, 
    breakable,                
    before skip=0pt,          
    after skip=0pt,
    halign=left
]
    \footnotesize\sffamily\textit{Evokes intrigue, curiosity, unknown, hidden truths.}

    \vspace{0.8em} 

    \textcolor{purple2}{\textbf{LLM Modifiers:}}
    \begin{itemize}[leftmargin=1.5em, itemsep=0.2em, topsep=0.2em, parsep=0em] 
        \item Focus on secrets, puzzles, and hidden truths. Introduce unanswered questions and obscured details.
        \item Describe enigmatic characters and subtle, confusing clues. Build anticipation without full revelation.
        \item Use evocative, atmospheric, and suggestive language. Employ fragmented sentences or rhetorical questions.
    \end{itemize}

    \vspace{0.8em} 

    \textcolor{purple2}{\textbf{General Sound Cues:}}
    Subtle, low-frequency drones; unsettling, barely audible hums. Distant, unresolved musical phrases. Isolated, unidentifiable creaks or whispers.

    \vspace{0.8em} 

    \textcolor{purple2}{\textbf{General Visual Cues:}}
    Dominant low-key lighting with deep, sprawling shadows. Obscured areas featuring mist, fog, or heavy rain. Frame shots with partially hidden objects or figures. Use slightly off-kilter or Dutch angles.

    \vspace{0.8em} 

    \textcolor{purple2}{\textbf{Intensity Specifics:}}

    \vspace{0.1em} 

    \begin{stagebox_2}[halign=left]
    \textcolor{purple1}{\textbf{Low Intensity:}}\\
    \textbf{LLM Keywords:} \textcolor{highlightpurple}{curious, peculiar, subtle hint} \\
    \textbf{Sound Keywords:} \textcolor{highlightpurple}{soft hum, distant rustle} \\
    \textbf{Visual Keywords:} \textcolor{highlightpurple}{subtle shadows, selective focus}      
    \end{stagebox_2}


    \begin{stagebox_2}[halign=left]
    \textcolor{purple1}{\textbf{Medium Intensity:}}\\
    \textbf{LLM Keywords:} \textcolor{highlightpurple}{enigmatic, unexplained, hidden, clues} \\
    \textbf{Sound Keywords:} \textcolor{highlightpurple}{eerie silence, unresolved chime, muffled whispers} \\
    \textbf{Visual Keywords:} \textcolor{highlightpurple}{obscured figures, moody lighting, half-glimpsed objects}
    \end{stagebox_2}


    \begin{stagebox_2}[halign=left]
    \textcolor{purple1}{\textbf{High Intensity:}}\\
    \textbf{LLM Keywords:} \textcolor{highlightpurple}{unfathomable, cryptic, dark truth, abyss} \\
    \textbf{Sound Keywords:} \textcolor{highlightpurple}{rising dissonant drones, piercing stings, fragmented whispers} \\
    \textbf{Visual Keywords:} \textcolor{highlightpurple}{impenetrable fog, silhouetted forms, extreme close-ups on obscured details}
\end{stagebox_2}
\end{tcolorbox}

\begin{figure*}[t]
    \centering
    \begin{subfigure}[b]{0.48\textwidth}
        \centering
        \includegraphics[width=\linewidth]{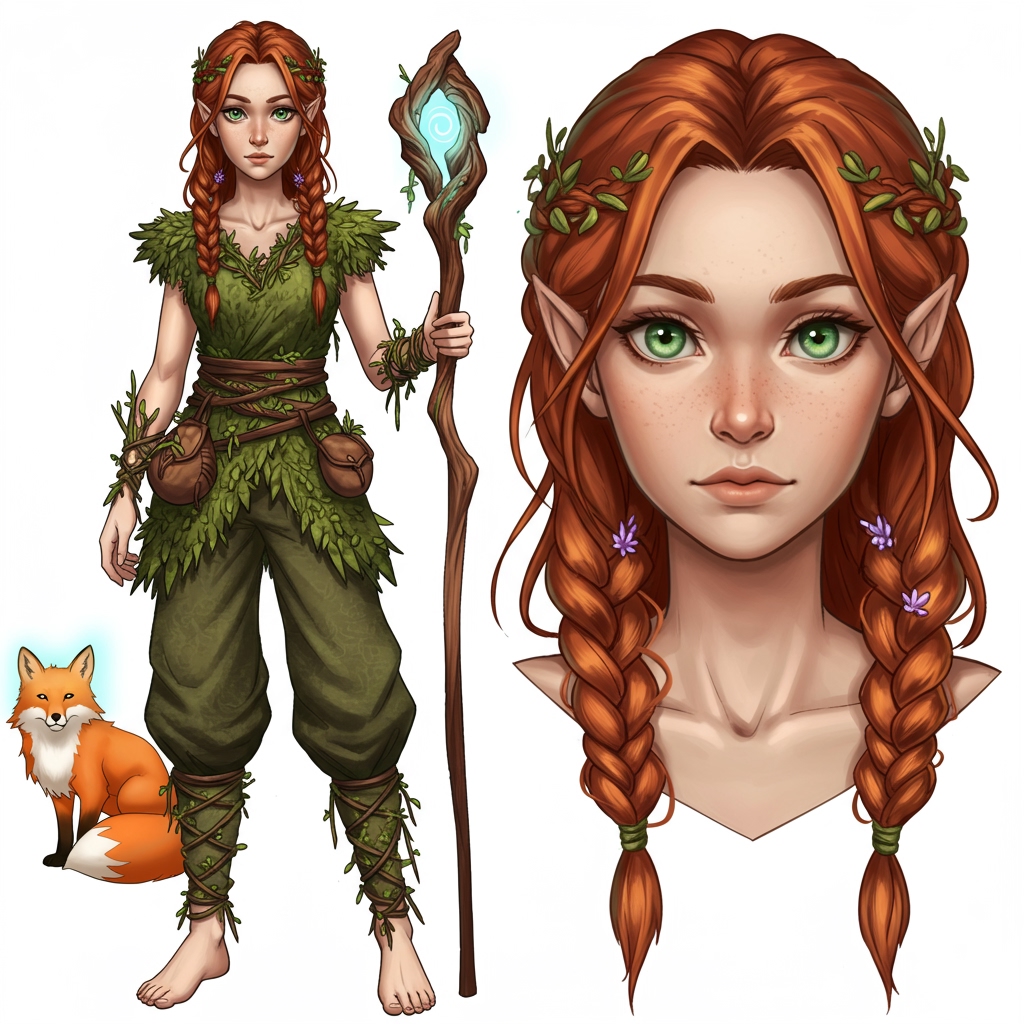}
        \caption{Character's visual}
    \end{subfigure}
    \hfill
    \begin{subfigure}[b]{0.48\textwidth}
        \centering
        \includegraphics[width=\linewidth]{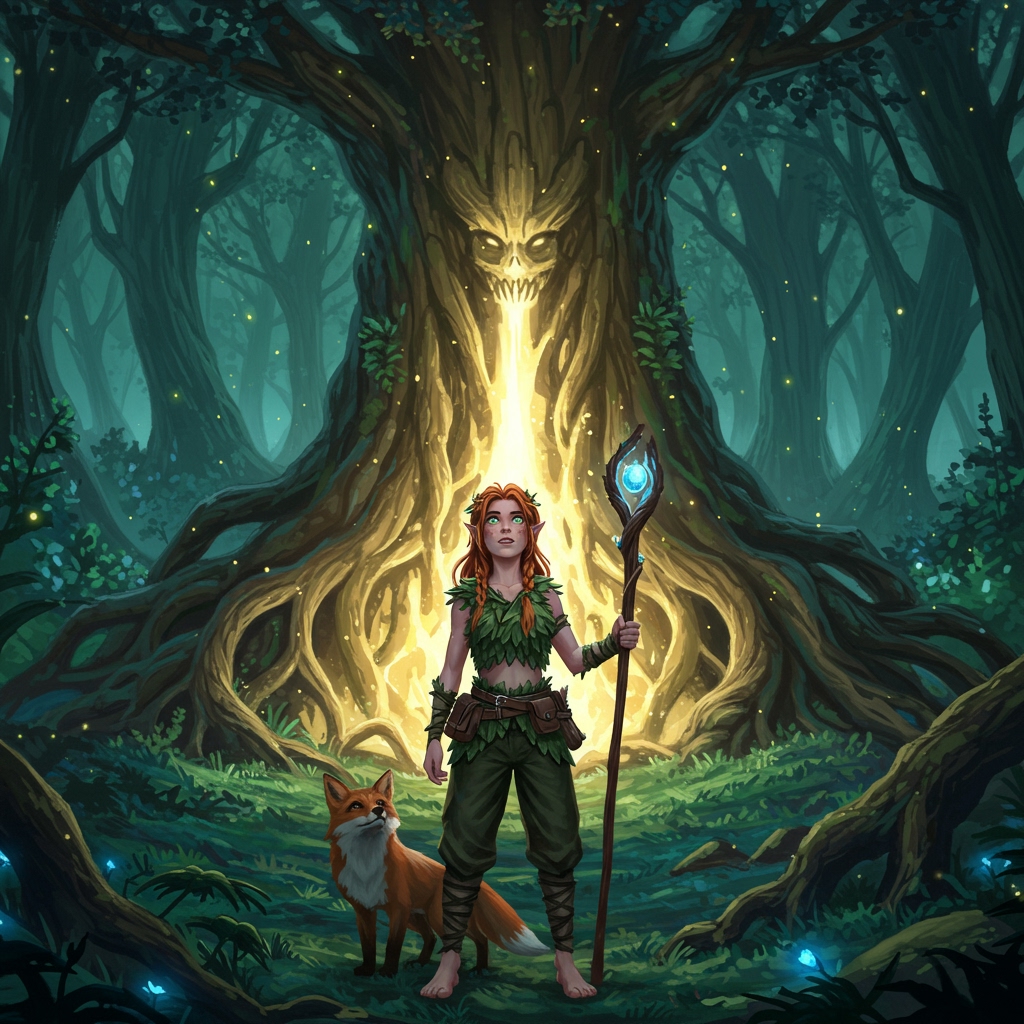}
        \caption{Scene image}
    \end{subfigure}
    \caption{Character grounding in Aether Weaver. (a) Shows the generated character's visual using available textual information. Then a vision model analyzes this image to extract and update the character's visual features with more detail. (b) The updated/extended character's visual features are then injected into the scene image generation prompt}
    \label{fig:character}
\end{figure*}

\subsection*{Multimodal Character Grounding}
Character profiles are managed through a combination of user inputs and image generation. Users can provide initial details like name, description, personality, and specific visual attributes (appearance, clothing, key features), or even upload a custom portrait image. If image generation model is tasked with creating a character's visual, it first generates a composite image (full-body and close-up) using all available textual information; then, a vision model analyzes this generated image (or a user-uploaded one if its textual visual details are generic) to extract and update the character's textual visual appearance details, consistent clothing, and key visual features fields, ensuring textual descriptions align with the visual representation. When new characters are discovered from the narrative, they undergo a similar process. For visualizing characters within a scene image, the system takes the general scene description and dynamically injects the detailed, up-to-date visual attributes (appearance, clothing, key features) from the relevant character profiles directly into the final image generation prompt if those characters are active in the scene, aiming for visual consistency with their established portraits and descriptions. \textbf{Figure \ref{fig:character}} illustrates an example of this character grounding process in Aether Weaver.

\newpage~
\subsection*{Scene Graph Example}
Below is an example of a scene graph generated by the Director, showcasing the nodes and edges that represent the current narrative state, including locations, characters, key objects, and events. The scene graph is designed to be dynamic, allowing for real-time updates as the narrative evolves.
\newpage~
\noindent

\newpage~
\begin{strip}
\begin{tcolorbox}[colback=background, colframe=purple, boxrule=1pt, arc=3mm, fontupper=\sffamily\color{white}, coltext=white, title empty, fonttitle=\bfseries\normalsize\sffamily, breakable, halign=left]
{\bfseries\large Nodes:}
\par

\begin{multicols}{2}

\begin{tcolorbox}[title=Locations:, colframe=purple, coltitle=white, colback=background, coltext=white, fonttitle=\bfseries\normalsize\sffamily, halign=left]
  \begin{tcolorbox}[fonttitle=\bfseries\small\sffamily, colframe=amber, colback=darkblue, coltext=white, title=Whispering Woods \texttt{(729e017b)}, fonttitle=\bfseries\normalsize\sffamily, halign=left]
    \textit{spatial\_description}: deep within, ancient trees, bioluminescent flora
  \end{tcolorbox}
  \begin{tcolorbox}[colframe=amber, colback=darkblue, coltext=white, title=Forgotten Shrine \texttt{(afe85846)}, fonttitle=\bfseries\normalsize\sffamily, halign=left]
    \textit{spatial\_description}: hidden in the eastern groves\\
    \textit{contains\_relic}: true\\
    \textit{relic\_power}: immense
  \end{tcolorbox}
\end{tcolorbox}

\begin{tcolorbox}[title=Characters Present:, colframe=purple, coltitle=white, colback=background, coltext=white, fonttitle=\bfseries\normalsize\sffamily, halign=left]
  \begin{tcolorbox}[colframe=skyblue, colback=darkblue, coltext=white, title=Elara \texttt{(de1cf932)}, fonttitle=\bfseries\normalsize\sffamily, halign=left]
    \textit{mood}: wonder and determination\\
    \textit{current\_location\_id}: cdbd9338-23a7-43e9-9623-937ba7a923d2
  \end{tcolorbox}
\end{tcolorbox}

\begin{tcolorbox}[title=Key Objects:, colframe=purple, coltitle=white, colback=background, coltext=white, fonttitle=\bfseries\normalsize\sffamily, halign=left]
  \begin{tcolorbox}[colframe=lightgreen, colback=darkblue, coltext=white, title=Elderwood \texttt{(cdbd9338)}, fonttitle=\bfseries\normalsize\sffamily, halign=left]
    \textit{description}: colossal, gnarled roots, breathtaking luminescence\\
    \textit{has\_spirit}: true
  \end{tcolorbox}
\end{tcolorbox}

\begin{tcolorbox}[title=Events:, colframe=purple, coltitle=white, colback=background, coltext=white, fonttitle=\bfseries\normalsize\sffamily, halign=left]
  \begin{tcolorbox}[colframe=salmonred, colback=darkblue, coltext=white, title=Spirit Revelation \texttt{(26117842)}, fonttitle=\bfseries\normalsize\sffamily, halign=left]
    \textit{Participants}: Elara, Elderwood\\
    \textit{Outcome}: \{``description'':``Spirit revealed growing blight and the existence/location of a powerful relic in a forgotten shrine.''\}\\
    \textit{Scene}: 1
  \end{tcolorbox}
\end{tcolorbox}
\end{multicols}
\hrulefill
\\
\vspace{1mm}
{\bfseries\large Edges:}
\par
\begin{itemize}[leftmargin=2em]
    \item \textcolor{skyblue}{Elara} \textcolor{purple1}{-(IS\_AT)$\rightarrow$}\textcolor{lightgreen}{Elderwood} \texttt{(ID: 625524a2)}
    \item \textcolor{lightgreen}{Elderwood} \textcolor{purple1}{-(INSIDE)$\rightarrow$}\textcolor{amber}{Whispering Woods} \texttt{(ID: aca05c17)}
    \item \textcolor{amber}{Forgotten Shrine} \textcolor{purple1}{-(INSIDE)$\rightarrow$}\textcolor{amber}{Whispering Woods} \texttt{(ID: 7ea31f73)}
\end{itemize}
\end{tcolorbox}
\end{strip}

\end{document}